\documentclass{article}
\usepackage{spconf,amsmath,graphicx}
\usepackage{hyperref}
\usepackage{siunitx}
\usepackage{xcolor,bm}
\usepackage{subcaption}
\usepackage{booktabs}
\usepackage{cite}

\title{Multispectral snapshot demosaicing via non-convex matrix completion}
\name{Giancarlo A. Antonucci$^*$, Simon Vary$^*$,  
David Humphreys\textsuperscript{$\ddagger$}, Robert A. Lamb\textsuperscript{$\ddagger$},
Jonathan Piper\textsuperscript{$\dagger$}, 
Jared Tanner$^{*,\sharp}$ \thanks{This publication is based on work partially supported by: the EPSRC Centre For Doctoral Training in Industrially Focused Mathematical Modelling (EP/L015803/1) in collaboration with DSTL, the EPSRC I-CASE studentship (voucher 15220165) in partnership with Leonardo, The Alan Turing Institute through EPSRC (EP/N510129/1) and the Turing Seed Funding grant SF019, Institut Henri Poincar\'e (UMS 839 CNRS-Sorbonne Universit\'e), and LabEx CARMIN (ANR-10-LABX-59-01).}}
\address{$*$ Mathematical Institute, University of Oxford, Oxford, OX2 6GG\\
$\sharp$ The Alan Turing Institute, NW1 2DB\\
$\dagger$ DSTL, Sensor Systems, Sensors and Countermeasures Department, Porton Down, Salisbury, SP4 0JQ\\
$\ddagger$ Leonardo MW Ltd., Edinburgh, EH5 2XS}

\begin{document}
\maketitle
\begin{abstract}
Snapshot mosaic multispectral imagery acquires an undersampled data cube by acquiring a single spectral measurement per spatial pixel. Sensors which acquire $p$ frequencies, therefore, suffer from severe $1/p$ undersampling of the full data cube.  We show that the missing entries can be accurately imputed using non-convex techniques from sparse approximation and matrix completion initialised with traditional demosaicing algorithms.  In particular, we observe the peak signal-to-noise ratio can typically be improved by \SIrange{2}{5}{\deci\bel} over current state-of-the-art methods when simulating a $p=16$ mosaic sensor measuring both high and low altitude urban and rural scenes as well as ground-based scenes.
\end{abstract}
\section{Introduction}
\label{sec:intro}
Multispectral imaging is the process of recording 2D arrays of information at multiple spectra (frequencies).  Having access to such a rich, three-dimensional data cube allows different materials to be distinguished due to their differing spectral emission profiles. As a result, multispectral imaging is used in applications ranging from landmine detection, precision agriculture, and medical diagnosis to name but a few of its application domains; for a partial survey see the January 2014 special issue of IEEE Signal Processing Magazine \cite{IEEESPM2014}. The increased sensor size and acquisition time are some of the central obstacles to the more widespread use of multispectral imagery.

Snapshot mosaic multispectral sensors allow for a compact video rate multispectral imaging by acquiring only a fraction of the multispectral cube. For example\footnote{We illustrate the architecture through the IMEC sensor, but note there are numerous similar sensors such as the S 137 system  by CUBERT.}, the IMEC SNm4x4 records 16 bands at a rate of 340 frames per second on a spatial two-dimensional $2048 \times 1088$ pixel domain by only acquiring a single spectrum per pixel; specifically this is achieved by tiling the two-dimensional domain by $4 \times 4$ pixel supercells where each supercell acquires the spectra independently. 

Herein we demonstrate the efficacy of multiple methods for interpolating the missing values in the above snapshot mosaic data cube by simulating the undersampling from complete three-dimensional data cubes provided by DSTL as well as AVIRIS \cite{Vane1993}, Stanford SCIEN \cite{Skauli2013}, Nascimento \cite{Nascimento2002}, Foster \cite{Foster2004}, IEEE GRSS Data Fusion Contest \cite{grss18}.  In addition to reviewing the existing state-of-the-art interpolation methods in Sec. \ref{sec:interp}, we demonstrate sparse approximation and matrix completion methods in Sec. \ref{sec:cs} and \ref{sec:mc} respectively, which we observe to substantially outperform the prior state-of-the-art. Over the above diverse data sets, we observe that non-convex compressed sensing and matrix completion methods initialised with traditional interpolations methods typically improve the peak signal-to-noise ratio by \SIrange{2}{5}{\deci\bel}, see Table \ref{tab:values}.

\section{Algorithms for demosaicing}
\label{sec:demosaic}
Demosaicing is the process by which the undersampled three-dimensional multispectral data cube has the missing entries approximated so as to simulate a full data acquisition.  While most three-dimensional interpolations methods would be directly applicable, we consider a few methods previously used by the multispectral community, such as direct interpolation as described in Sec. \ref{sec:interp} 
as well as sparse approximation regularisation methods in Sec. \ref{sec:cs} and low-rank structure as presented in Sec. \ref{sec:mc}.

\subsection{Direct interpolation methods}
\label{sec:interp}
Brauers et al. \cite{Brauers2006} developed methods to estimate the missing values in the multispectral cube based on extending a spatial bilinear interpolation of the measured values to include any spectral correlation.  The weighted bilinear interpolation (WB) for the $4 \times 4$ pixel regular mosaic filter follows by padding the missing entries with zeros and convolving with the cartesian product of a discrete seven-pixel width filter $\frac{1}{4}[1\; 2\; 3\; 4\; 3\; 2\; 1]$.  Then, the spectral correlation is included in the spectral difference (SD) method by a) taking the output of WB to independently compute, for each band, say $k$, the difference between the values of the measured pixels for spectrum $k$ and the WB interpolated values of every other band restricted to the support of the measured pixels of spectrum $k$, then b) applying WB to these spectral differences c) to form an approximation of the full spectrum $k$ by adding the output of step (b) to the difference with $l$ at the location of the measured pixels for spectrum $l$.

Mihoubi et al. \cite{Mihoubi2015} extended the SD approach to consider alternative ways to build correlations between the bands. In intensity difference (ID) they build spectral correlation by first constructing a spatial intensity map whose value at a pixel is the measurement for whichever spectra was measured at that spatial location, then this intensity map is averaged using a weighting based on the number of pixels per spectra contained in the averaging width. See Sec.\ref{sec:setup} for details on the choice of averaging used here and Mihoubi et al. \cite{Mihoubi2015} for a more general discussion. Hence, the difference between this averaged intensity map and the measurements for each spectrum is computed, the unknown values zero padded, and each band averaged such as in WB.

Interpolation methods designed in transform domains have been considered by Miao et al. \cite{Miao2006} in the binary tree-based edge-sensing (BTES) method, which has the additional benefit of allowing for variable sampling densities per frequency band. However, we observe it is inferior to SD and ID described above in the setting of snapshot imaging. Pseudo-panchromatic image difference (PPID) \cite{Mihoubi2017} builds upon BTES and ID. However, due to the applicability of PPID to only some specific mosaic arrangements we leave comparison with our algorithms for a later time.

\subsection{Sparse approximation and compressed sensing inpainting}
\label{sec:cs}
Sparse approximation inpainting allows one to easily consider the interpolation of the under-complete snapshot data cube in transforms more general than the linear interpolation of (WB).  In particular, one can assume that the image is well approximated by a sparse representation in a suitable image domain and exploit this structure to reconstruct it from undersampled measurements \cite{Kutyniok2013, Dong2012, Elad2005}, e.g. by solving
\begin{equation}
    \min_{x} \|y - P_\Omega \Psi^{-1} x\|_2 \,, \quad \text{s.t.}\ \| x \|_0 \leq k \,,
    \label{eq:problem_cs}
\end{equation}
where $\Psi$ represents the transform in which the data is known to be compressible and $y$ is the full data cube projected by $P_{\Omega}$ to the undersampled locations.  Degraux et al. \cite{Degraux2015} apply this model to a reconstruction of multispectral imagery acquired by mosaic snapshot cameras. 

The primary challenge lies in two aspects: (i) the significant subsampling ratio of $1/K$, where $K$ is the number of spectral bands, and (ii) the selection of the suitable transform $\Psi$. The first problem can be overcome by initialising the state-of-the-art sparse approximation algorithms for solving \eqref{eq:problem_cs} with sufficiently accurate initial estimates, such as those from the classical interpolation methods described in Sec. \ref{sec:interp}. As it was pointed out in \cite{Degraux2015}, we find that, for natural scenes captured by snapshot multispectral imaging, a Kronecker product of 2D wavelet transform spatially and the discrete cosine transform for the spectral dimension is an effective choice for the representation $\Psi$. In particular, the 2D wavelet transform spatially includes elements of nearly global support to allow broad correlations as well as local elements to express fine detail and the discrete cosine transform models the slowly varying values in the spectral dimension.

\subsection{Matrix completion}
\label{sec:mc}
Rather than using local correlations, matrix (and tensor) completion exploit the correlation in the data cube through a low-rank structure, e.g. by solving
\begin{equation}
    \min_{X} \|y - P_\Omega X\|_2 \,, \quad \text{s.t.}\ \mathrm{rank}(X) \leq r \,, \label{eq:problem_mc}
\end{equation}
where $y$ is the observed data, $X$ is a matrix corresponding to an unfolding of the complete three-dimensional data cube, and $P_\Omega$ is a restriction to the measured values as before.

Although the low-rank matrix completion problem is NP-hard in general (see \cite{mc_survey} for a recent survey) there is a number of computationally fast solvers for the problem with provable convergence guarantees \cite{Wen2012, Tanner2016, Blanchard2015}. In fact, matrix completion has been previously applied to the reconstruction of subsampled multispectral imagery \cite{Gelvez2015} by the Coded Aperture Snapshot Spectral Imager (CASSI) \cite{Gehm2007}. Here we show that matrix completion can be used also in the case of a more severe subsampling by snapshot mosaic camera designs if provided with suitable initialisation.

While there are many non-convex methods for matrix completion, we showcase two exemplary cases but expect that other non-convex methods would perform similarly well.
We apply conjugate gradient iterative hard thresholding (CGIHT) \cite{Blanchard2015} and alternating steepest descent (ASD) \cite{Tanner2016} to solve \eqref{eq:problem_mc}, providing them with an initial guess from either SD or ID. 
We show that both CGIHT and ASD can improve on the classical interpolation methods.  This differs substantially from prior work both in terms of using more recent algorithms for matrix completion which have been shown to be more effective and initialising them with prior state-of-the-art interpolation methods, and moreover in that unlike \cite{Gelvez2015} which treat each spectral band separately with $30\%$ undersampling, we vectorise the spatial dimensions to create a matrix of size $2,228,224$ by $16$ with $1/16$ undersampling.  We observe that this unfolding which allows full correlation between the spectral information is particularly effective, often resulting in reconstructions which are visually indistinguishable from fully acquired data.

\section{Numerical simulations}
\label{sec:numerics}
In this section, we show and explain the numerical results obtained by applying the methods discussed above.

\subsection{Data sets}
\label{sec:data}
We consider the efficacy of the algorithms for demosaicing on the following data sets:
\begin{itemize}
    \item High altitude airborne images from the AVIRIS \cite{Vane1993} and 2018 IEEE GRSS Data Fusion contest \cite{grss18}. AVIRIS  line-scan captures $224$ spectral bands in the \SIrange{380}{2500}{\nano\metre} and the GRSS images have $48$ spectral bands in the range of \SIrange{380}{1050}{\nano\metre}.
    \item Low altitute airborne images acquired at DSTL Porton Down, in August 2014, from which we selected $10$ representative radiance images of fields from a HySpex VNIR-1600 line-scan camera in the range \SIrange{400}{1000}{\nano\metre}.
    \item Ground-based images from the Stanford SCIEN \cite{Skauli2013}, Nascimento \cite{Nascimento2002} and Foster \cite{Foster2004}. The Stanford SCIEN images come from the line-scan HySpex VNIR-1600 camera.
\end{itemize}

We processed these data sets to simulate the spectrum measures by the IMEC SNm4x4 snapshot sensor with access to the complete data cube.  Then, we undersampled the data cube following the sensor sampling pattern and the following simulations conducted. 

\subsection{Simulation setup}
\label{sec:setup}
We implement and test recovery by two interpolation methods ID and SD, two matrix completion methods ASD and CGIHT and a compressed sensing version of CGIHT with a sparsifying transform as a Kronecker products of 2D Daubechies wavelets (W2) in the spatial and 1D Discrete Cosine Transform  (DCT) in the spectral domain which we reference as W2$\times$DCT.

The iterative algorithms are terminated once the error in iteration $t$ is $\| P_\Omega X^{(t)} - y \|_2 / \|y\|_2 \leq 10^{-7}$ or at the $500^{th}$ iteration. 
\begin{figure}[t!]
\centering
\subcaptionbox{
    PSNR throughout spectral bands.
    \medskip
    \label{fig:1a}
}{
    \begin{minipage}[b]{1.0\linewidth}
        \includegraphics[width=8.7cm]{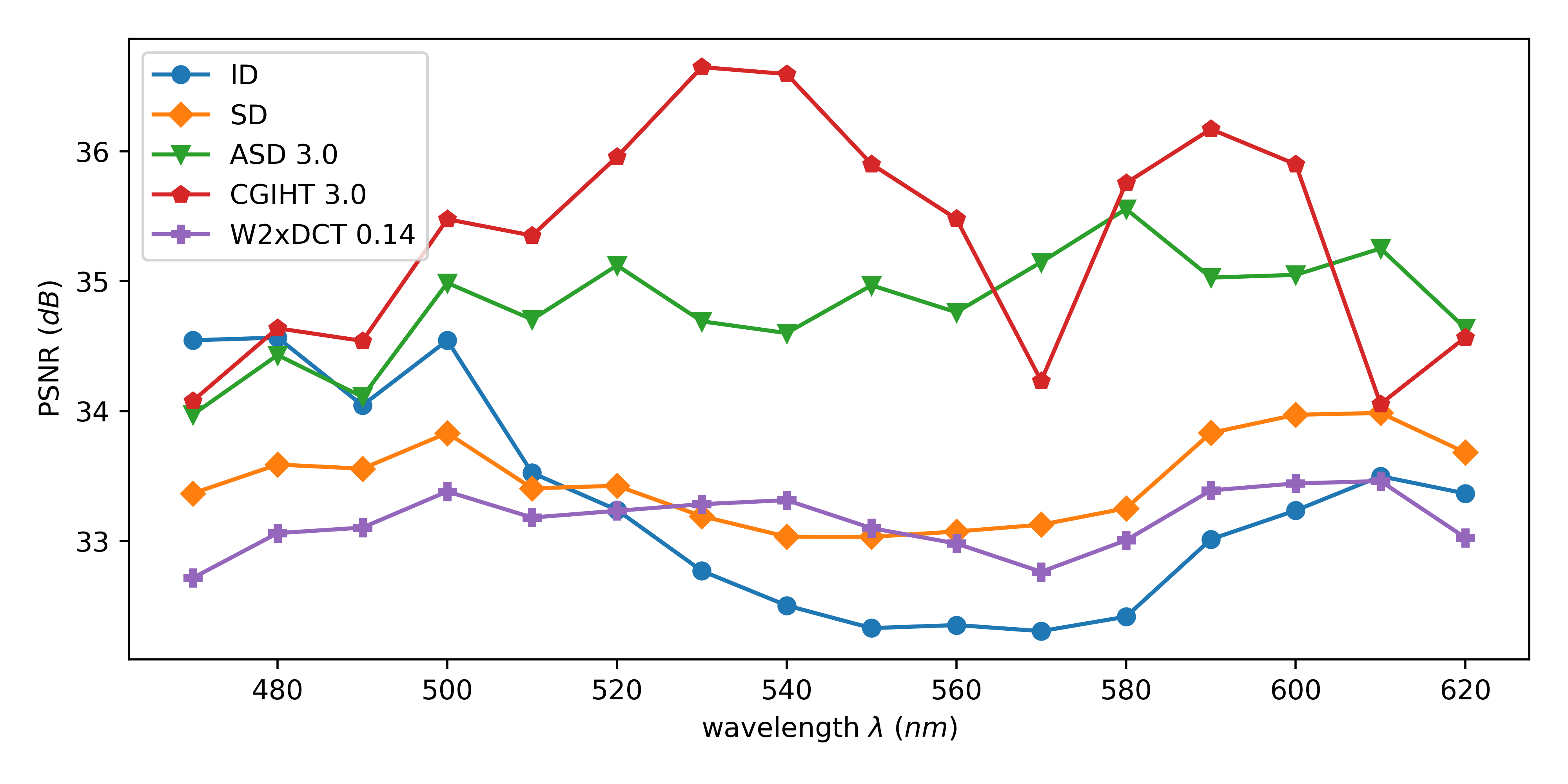}
    \end{minipage}
}
\subcaptionbox{
    W2$\times$DCT [\SI{33.2}{\deci\bel}].
    \label{fig:1b}
}{
    \begin{minipage}[b]{.45\linewidth}
        \includegraphics[width=4.2cm]{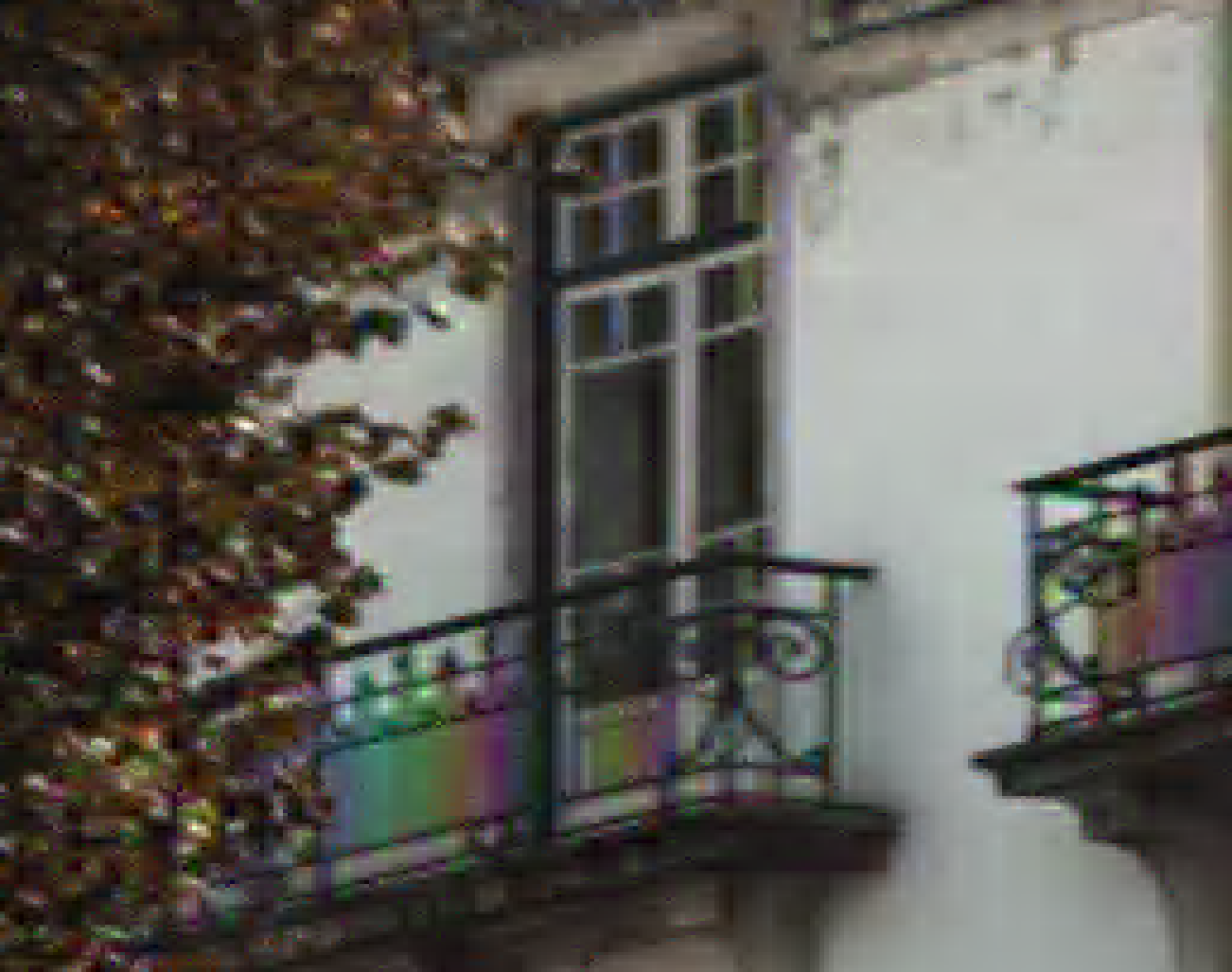}
    \end{minipage}
}
\hfill
\subcaptionbox{
    CGIHT [\SI{35.3}{\deci\bel}].
    \label{fig:1c}
}{
    \begin{minipage}[b]{0.45\linewidth}
        \includegraphics[width=4.2cm]{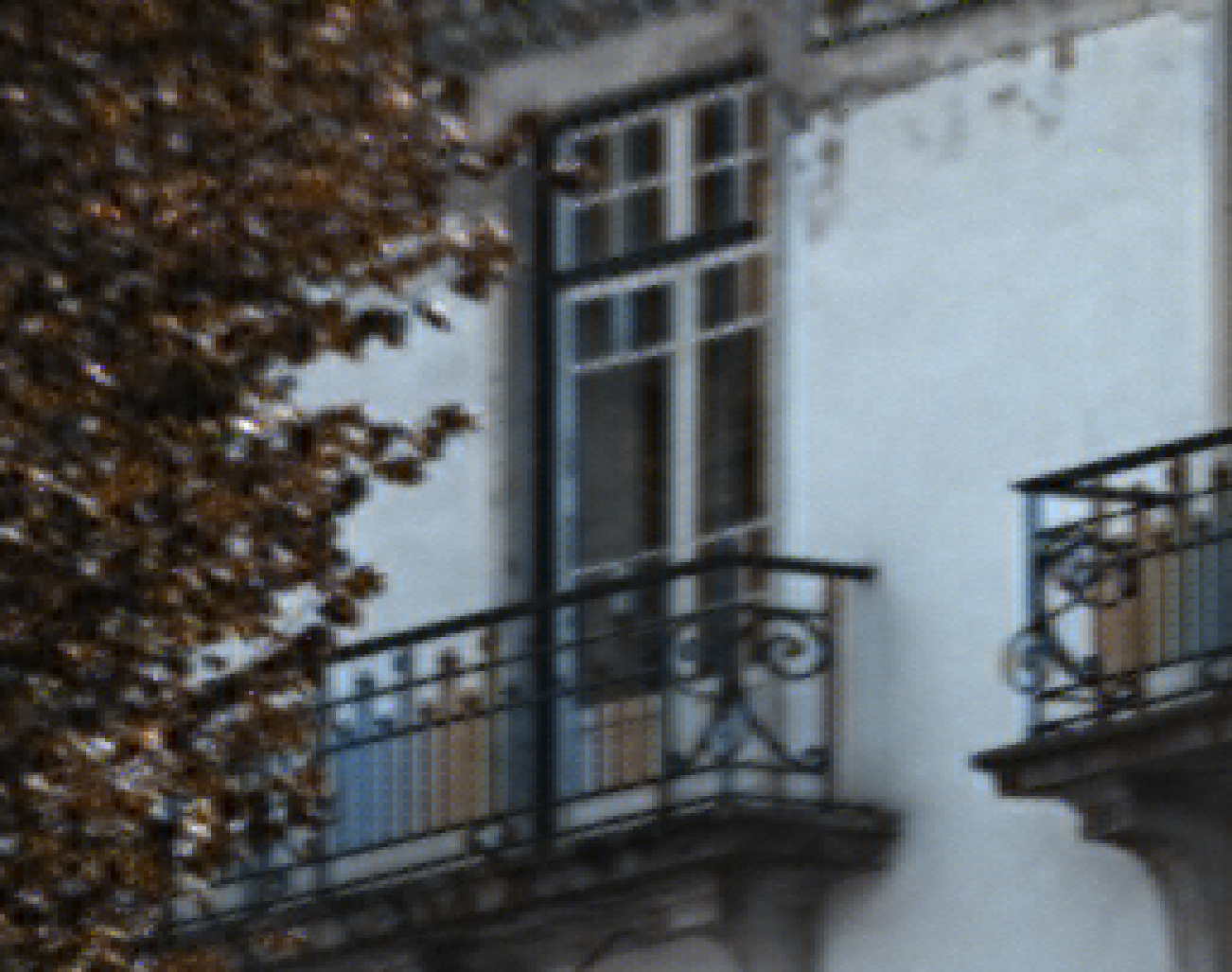}
    \end{minipage}
}
\caption{Results for reconstruction of \texttt{N06}. W2$\times$DCT (CS) and CGIHT (MC) initialised from SD. Wavelet based CS method smooths out the image while CGIHT MC is able to better preserve sharp edges.}
\label{fig:results1}
\end{figure}

We report the quality of an image approximated by demosaicing by the peak signal-to-noise ratio (PSNR), defined as the log of the ratio between the maximum possible power of (the slide of) an image and the power of corrupting noise that affects the fidelity of its representation, computed in terms of the average squared difference (or mean squared error, MSE) between the reference image and its reconstruction: 
\begin{equation}
    \text{PSNR}_k = 10 \log_{10} \left(\frac{(\max_{p \in \mathcal{P}} I_p^k)^2}{\frac{1}{|\mathcal{P}|}\sum_{p \in \mathcal{P}} (I_p^k - \hat{I}_p^k)^2}\right) \,,
\end{equation}
where $I^k$ and $\hat{I}^k$ are the $k$-th band slices of the reference cube and the reconstruction, respectively, and $\mathcal{P}$ denotes the set of all pixels.

We also employ the structural similarity (SSIM) index \cite{Zhou2004}, which is a decimal value between $-1$ and $1$, with $1$ being reachable only in the case of two identical sets of data. SSIM is a perception-based model that considers image degradation as perceived change in structural information, while also incorporating important perceptual phenomena, including both luminance masking and contrast masking terms.

\subsection{Results}
\label{sec:results}
Figure \ref{fig:1a} shows the PSNR of each spectrum given its band centre, for a sample image from Nascimento \cite{Nascimento2002}, for the classical interpolation methods SD and ID as well as \hbox{the compressed sensing} (CS) and matrix completion techniques initialised with SD. Notice that, except for the first band, the matrix completion algorithms outperform SD and ID. On the other hand, the compressed sensing approach does not improve on the interpolation methods. In particular, note the overall incorrect contrast level resulting in yellowing of Figure \ref{fig:1b}. Moreover, we lose the sharpness of the edges in the balcony when employing CS W2$\times$DCT (Figure \ref{fig:1b}), while we recover it with the matrix completion variant of CGIHT (Figure \ref{fig:1c}).

\begin{figure}[t]
\subcaptionbox{
    \texttt{D2301}, ID [\SI{37.9}{\deci\bel} PSNR and SSIM of 0.99992].
    \label{fig:2c}
}{
    \begin{minipage}[b]{.48\linewidth}
        \centering
        \includegraphics[width=4.0cm]{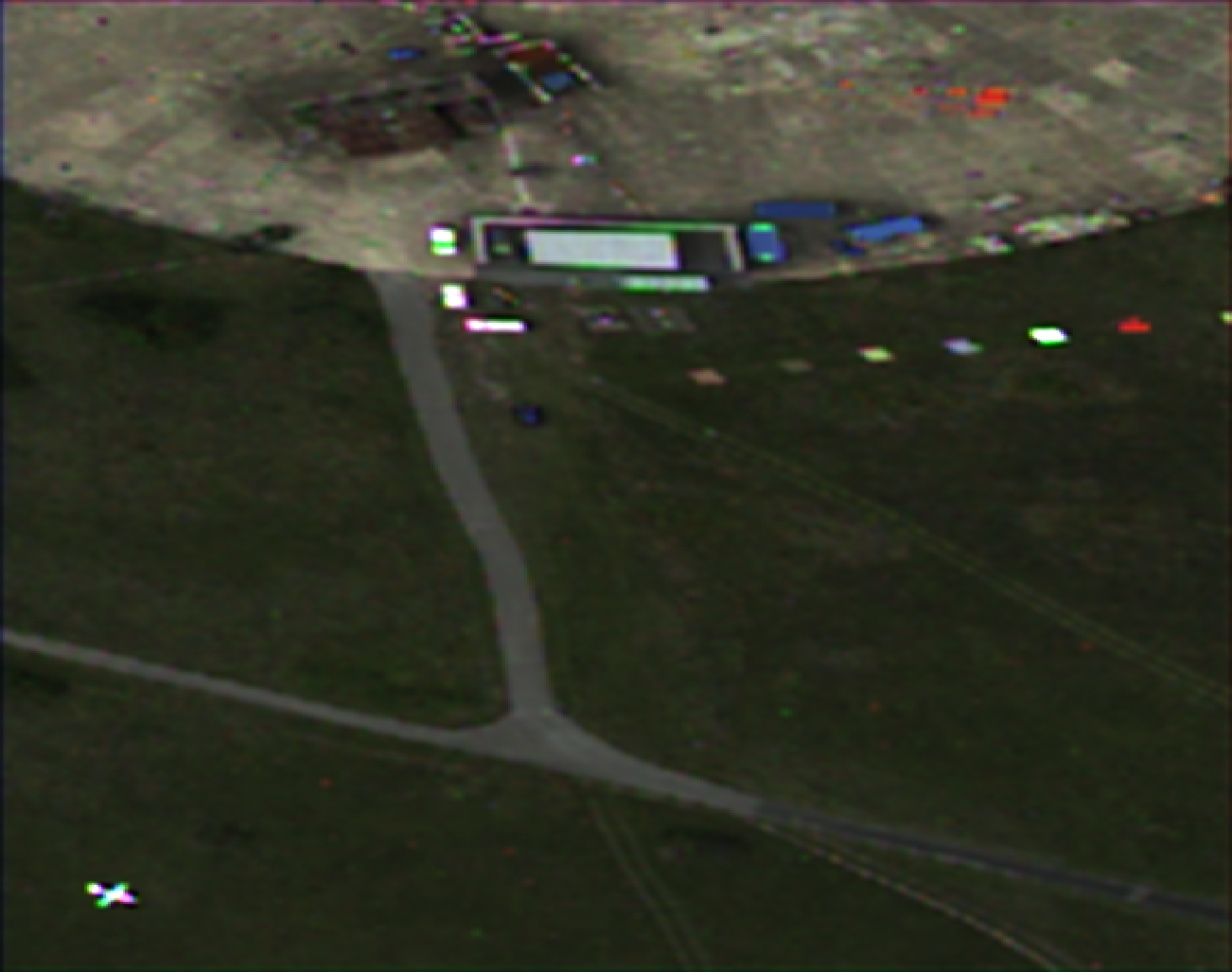}
    \end{minipage}
    \hfill
    \begin{minipage}[b]{0.48\linewidth}
        \centering
        \includegraphics[width=4.31cm]{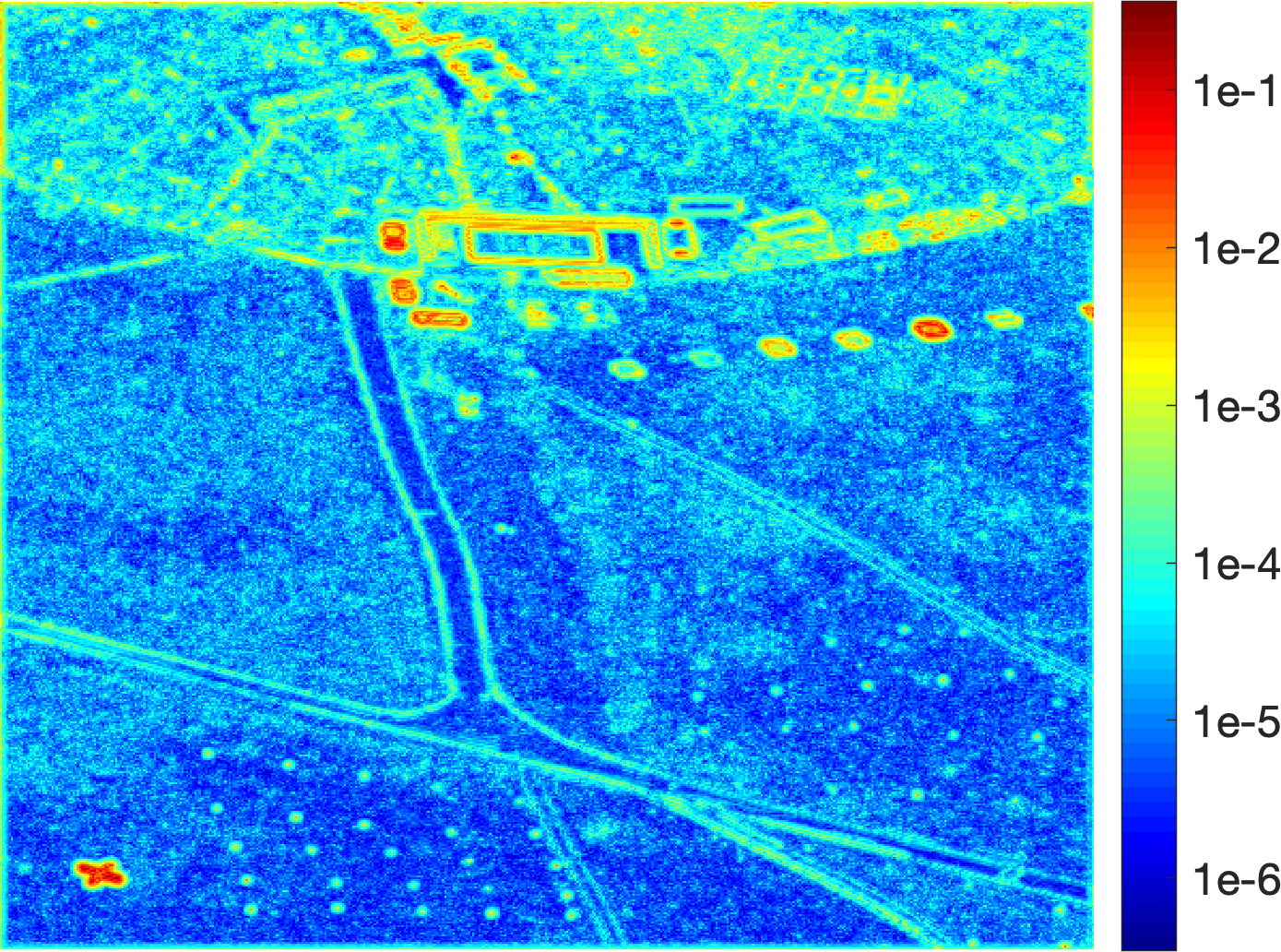}
    \end{minipage}
}
\subcaptionbox{
    \texttt{F06}, W2$\times$DCT from SD [\SI{39.3}{\deci\bel} PSNR and SSIM of 0.99955].
    \label{fig:2b}
}{
    \begin{minipage}[b]{.48\linewidth}
        \centering
        \includegraphics[width=4.0cm]{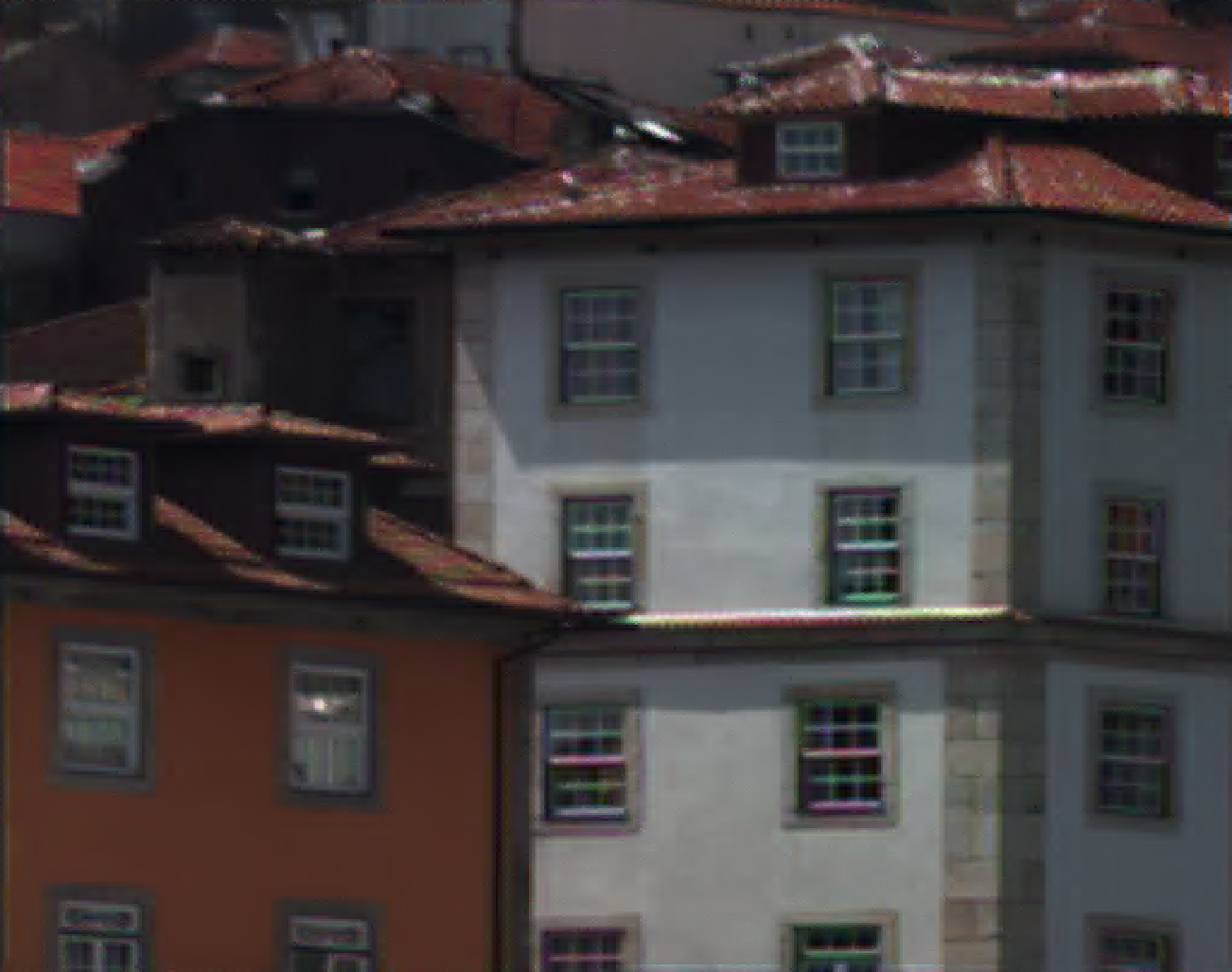}
    \end{minipage}
    \hfill
    \begin{minipage}[b]{0.48\linewidth}
        \centering
        \includegraphics[width=4.31cm]{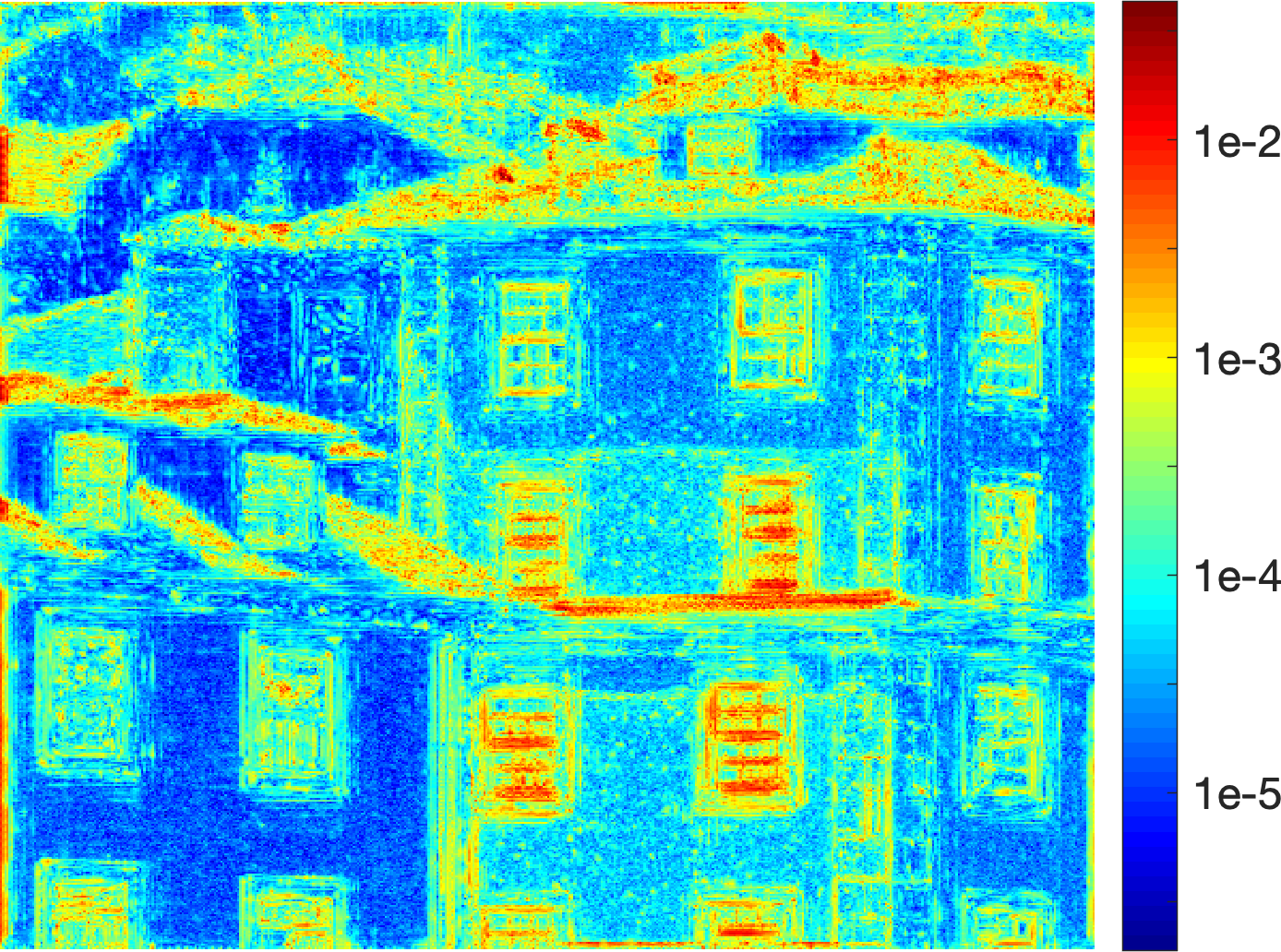}
    \end{minipage}
}
\subcaptionbox{
    \texttt{GD}, CGIHT from SD [\SI{36.2}{\deci\bel} PSNR and SSIM of 0.99988].
    \label{fig:2a}
}{
    \begin{minipage}[b]{.48\linewidth}
        \centering
        \includegraphics[width=4.0cm]{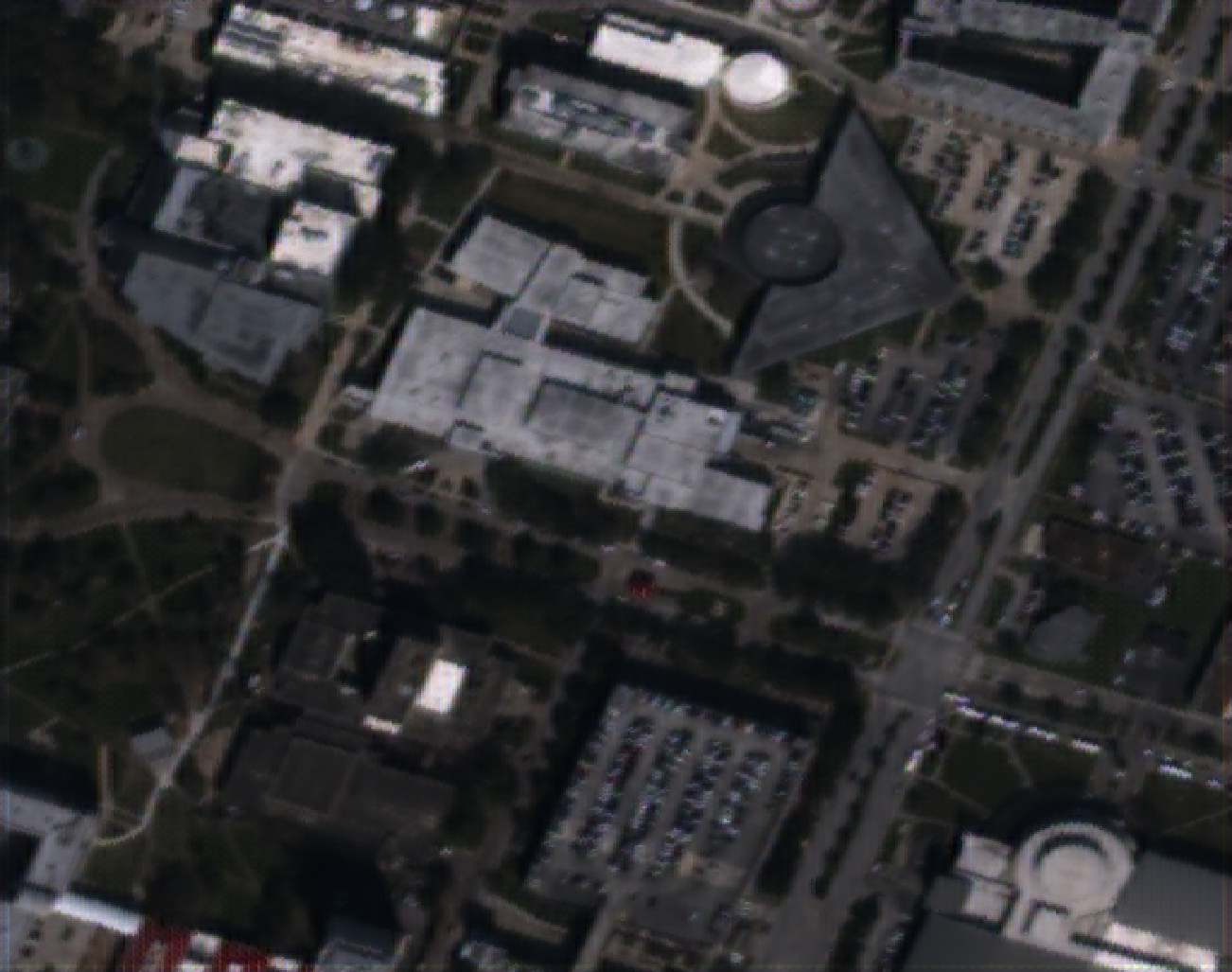}
    \end{minipage}
    \hfill
    \begin{minipage}[b]{0.48\linewidth}
        \centering
        \includegraphics[width=4.31cm]{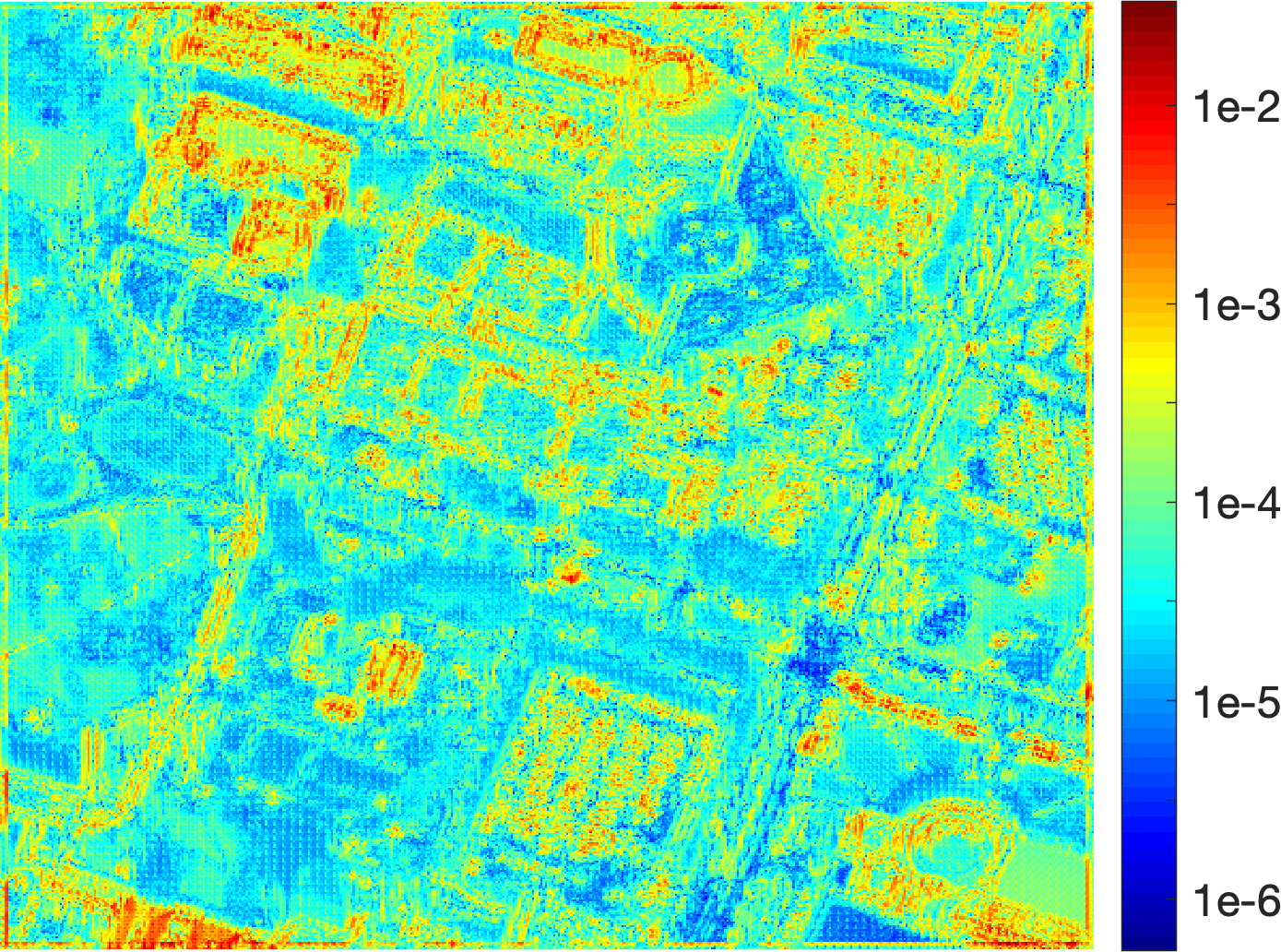}
    \end{minipage}
}
\caption{Left: Colour renderings of image reconstructions. Right: MSE of reconstructions (log-scale).}
\label{fig:results2}
\vspace{-1em}
\end{figure}

To further emphasise how the different algorithms differ from each other we show the reconstructions PSNR from the corresponding reference images in Figure \ref{fig:results2}. Note in Figure \ref{fig:2a} how ID accurately reconstructs the field image, taking into account the spectral correlation between the bands, but smooths the sharp details. The compressed sensing approach does a better job in identifying the edges (Figure \ref{fig:2b}), but suffers from the same problem overall. On the other hand, the matrix completion CGIHT outperforms the other methods due to the presence of field-like uniformities.
\begin{table}[t!]
    \centering
    \resizebox{\columnwidth}{!}{
    \begin{tabular}{|c|c|c|c|c|c|}
        \hline
        IMAGE & \multicolumn{2}{c|}{INIT} &ASD &CGIHT &W2$\times$DCT \\
        \hline
        \texttt{D0201}%
        & ID &$34.5\pm0.8$ &$36.3\pm1.1$ &$37.9\pm1.1$ &$34.3\pm0.5$ \\
        & SD & $35.9\pm0.7$ &$38.1\pm0.9$ &$\bm{39.5\pm1.2}$ &$35.8\pm0.6$ \\
        \hline
        \texttt{D0301}%
        & ID &$36.0\pm1.8$ &$38.3\pm2.2$ &$39.2\pm1.7$ &$35.1\pm1.1$ \\
        & SD &$37.3\pm1.3$ &$39.7\pm1.4$ &$\bm{40.5\pm1.5}$ &$36.3\pm1.0$ \\
        \hline
        \texttt{D0303}%
        & ID &$39.1\pm1.0$ &$41.2\pm1.2$ &$43.1\pm1.4$ &$38.8\pm0.7$ \\
        & SD &$40.8\pm0.8$ &$43.4\pm1.1$ &$\bm{45.5\pm1.3}$ &$40.5\pm0.7$ \\
        \hline
        \texttt{D0307}%
        & ID &$34.1\pm1.8$ &$36.1\pm2.3$ &$36.9\pm2.0$ &$33.2\pm1.1$ \\
        & SD &$35.2\pm1.1$ &$37.2\pm1.2$ &$\bm{37.8\pm1.5}$ &$34.3\pm0.8$ \\
        \hline
        \texttt{D2301}%
        & ID &$37.9\pm0.3$ &$39.4\pm0.7$ &$41.1\pm1.3$ &$37.8\pm0.3$ \\
        & SD &$39.5\pm0.3$ &$41.5\pm1.0$ &$\bm{43.6\pm1.5}$ &$39.7\pm0.4$ \\
        \hline
        \texttt{AvLF}%
        & ID &$33.7\pm0.7$ &$35.6\pm1.2$ &$36.7\pm2.3$ &$33.5\pm0.7$ \\
        & SD &$35.3\pm0.7$ &$37.8\pm1.6$ &$\bm{39.3\pm2.6}$ &$34.7\pm0.6$ \\
        \hline
        \texttt{StCh}%
        & ID &$36.2\pm1.0$ &$36.7\pm1.7$ &$36.6\pm1.8$ &$36.0\pm1.0$ \\
        & SD &$37.3\pm1.1$ &$\bm{37.5\pm1.0}$ &$37.2\pm0.9$ &$37.4\pm1.0$ \\
        \hline
        \texttt{N04}%
        & ID &$37.3\pm3.1$ &$\bm{38.6\pm2.6}$ &$38.2\pm2.8$ &$35.3\pm1.9$ \\
        & SD &$35.6\pm0.8$ &$36.9\pm0.8$ &$35.4\pm2.2$ &$34.9\pm0.9$ \\
        \hline
        \texttt{N06}%
        & ID &$33.3\pm0.8$ &$34.5\pm0.6$ &$35.2\pm0.5$ &$32.6\pm0.4$ \\
        & SD &$33.5\pm0.3$ &$34.8\pm0.4$ &$\bm{35.3\pm0.8}$ &$33.2\pm0.2$ \\
        \hline
        \texttt{N08}%
        & ID &$33.5\pm0.4$ &$34.4\pm0.6$ &$\bm{35.6\pm1.0}$ &$32.7\pm0.2$ \\
        & SD &$33.2\pm0.4$ &$34.4\pm0.7$ &$35.5\pm1.3$ &$32.9\pm0.3$ \\
        \hline
        \texttt{F05}%
        & ID &$36.1\pm2.4$ &$\bm{36.6\pm1.9}$ &$36.5\pm1.9$ &$34.0\pm1.4$ \\
        & SD &$35.0\pm0.4$ &$36.1\pm0.8$ &$35.7\pm1.6$ &$34.1\pm0.7$ \\
        \hline
        \texttt{F06}%
        & ID &$40.0\pm0.9$ &$\bm{40.2\pm0.8}$ &$39.9\pm0.9$ &$38.6\pm0.7$ \\
        & SD &$38.9\pm0.4$ &$39.6\pm0.9$ &$39.3\pm0.9$ &$39.1\pm0.5$ \\
        \hline
        \texttt{F07}%
        & ID &$34.9\pm1.2$ &$\bm{36.3\pm1.5}$ &$36.1\pm1.6$ &$34.4\pm0.9$ \\
        & SD &$34.6\pm0.8$ &$36.3\pm1.2$ &$35.1\pm1.8$ &$34.6\pm0.8$ \\
        \hline
        \texttt{GB}%
        & ID &$34.3\pm1.1$ &$36.0\pm0.9$ &$36.7\pm1.1$ &$33.3\pm0.5$ \\
        & SD &$34.9\pm0.3$ &$37.2\pm0.9$ &$\bm{37.5\pm1.5}$ &$34.7\pm0.7$ \\
        \hline
        \texttt{GD}%
        & ID &$34.1\pm1.4$ &$35.8\pm1.1$ &$36.2\pm1.2$ &$33.0\pm0.5$ \\
        & SD &$34.6\pm0.4$ &$\bm{36.9\pm0.9}$ &$36.6\pm1.1$ &$34.3\pm0.7$ \\
        \hline
        \texttt{GR}%
        & ID &$35.3\pm1.2$ &$37.0\pm1.0$ &$37.3\pm1.2$ &$34.4\pm0.5$ \\
        & SD &$36.0\pm0.4$ &$38.0\pm0.9$ &$\bm{38.1\pm1.7}$ &$35.7\pm0.6$ \\
        \hline
    \end{tabular}
    }
    \caption{Average PSNR over the 16 bands, with standard deviations. The best results for each image are highlighted in bold.}
    \label{tab:values}
    \vspace{-1em}
\end{table}
Finally, as shown in Table \ref{tab:values}, in the majority of cases the best performing algorithms are the matrix completion CGIHT and ASD initialised with SD, with improvements over both SD and ID from \SIrange{2}{5}{\deci\bel}. \texttt{StCh} from Stanford SCIEN \cite{Skauli2013} seems to be the only outlier, with an improvement of just \SI{0.2}{\deci\bel}.

Being directly related to the number of bands, the rank of the spectral unfolding seems to be effective in capturing the spectral information of the analysed datasets. Our results suggest a high correlation between the frequency bands and a low-rank structure of the spectral unfolding of our images, in which most of the information is contained in the first 3 singular values of the spectral unfolding.

\section{Conclusions}
\label{sec:summary}
We provide a numerical comparison of multispectral demosaicing by traditional interpolation, sparse approximation and matrix completion methods. Our experiments demonstrate that non-convex matrix completion typically improves reconstruction by \SIrange{2}{5}{\deci\bel} over the current state-of-the-art methods. This differs substantially from prior work in terms of employing matrix completion on the spectral unfolding of the image in the context of demosaicing, initialising it with classical interpolation methods and using more recent non-convex matrix completion algorithms.

\bibliographystyle{IEEEbib}
\bibliography{refs}

\end{document}